\documentclass{article}

 \usepackage[main, final]{neurips_2026}

\usepackage[utf8]{inputenc} 
\usepackage[T1]{fontenc}    
\usepackage{hyperref}       
\usepackage{url}            
\usepackage{booktabs}       
\usepackage{amsfonts}       
\usepackage{nicefrac}       
\usepackage{microtype}      
\usepackage{xcolor}         
\usepackage{amsmath}
\usepackage{float}
\usepackage{graphicx}
\usepackage{multirow}
\usepackage{caption}
\usepackage{pifont}
\usepackage{subcaption}

\newcommand{\nop}[1]{}
\newcommand{\from}[2]{{\bf [{\sc from #1:} #2]}}

\title{Dual Dimensionality for Local and Global Attention}

\author{
Zhiyuan Wang \\
UC Santa Barbara\\
\texttt{zwang796@ucsb.edu} \\
\And
Xuan Luo \\
UC Santa Barbara\\
\texttt{xuan\_luo@cs.ucsb.edu} \\
\AND
Sirui Zeng \\
UC Santa Barbara\\
\texttt{sirui\_zeng@ucsb.edu} \\
\And
Xifeng Yan \\
UC Santa Barbara\\
\texttt{xyan@cs.ucsb.edu}
}

\begin{document}

\maketitle

\begin{abstract}
Decoder-only Transformers compute attention over the KV cache of preceding tokens. Keys and Values are typically represented with the same dimensionality, regardless of their distance from the prediction target. In natural language, however, the next word is most strongly influenced by the immediately preceding tokens. We hypothesize that local and distant tokens impose asymmetric demands on representational capacity: local tokens are more critical for predicting immediate outputs and thus require richer representations, whereas distant tokens primarily serve as long-range memory, for which lower-dimensional representations may suffice. We formalize this idea as Distance-Adaptive Representation (DAR), which preserves full-dimensional representations within a local context window while assigning reduced-dimensional representations (e.g. 1/2 or 1/4 of the original dimensionality) to tokens beyond that window. Across multiple pretraining scales (70M to 410M parameters), as well as continued supervised fine-tuning on a 1B-scale model, this approach closely matches the performance of full-dimensional baselines. In contrast, uniformly reducing dimensionality across all token positions leads to worse performance. These results challenge the common assumption that key and value dimensionality should be uniform across token positions. We further develop an efficient realization of DAR that achieves model quality comparable to the MLA baseline while reducing the average KV-cache size by 68.8\% at 8K context and achieving up to a $1.9\times$ decoding speedup at 128K context on the evaluated models. Our findings suggest a new direction for designing attention architectures that allocate representational capacity according to token distance to reduce KV-cache during inference.

\end{abstract}

\section{Introduction}
\label{sec:intro}

The success of Transformer-based language models is largely attributed to the self-attention mechanism~\cite{vaswani2017attention}, which allows each token to attend to all preceding context. In standard Transformers, every previous token contributes key and value states of the same dimensionality, regardless of its distance from the current prediction target. This reflects an implicit architectural assumption that the representational capacity required of past tokens does not depend on how far they are from the token being predicted.

We revisit this assumption motivated by a simple observation about natural language. When producing a sequence of words, the most recent context has direct effects on the next word, such as avoiding immediate repetition, following local grammatical rules, and keeping sentiment consistent, while more distant context provides long-range memory and context. This asymmetry suggests that local and distant tokens may contribute different kinds of information to next-token prediction. Distant tokens may primarily provide long-range information that can be represented with lower dimensionality.

\begin{figure}[t]
    \centering
    \includegraphics[
        width=0.6\linewidth,
        trim={10 10 10 10},
        clip
        ]{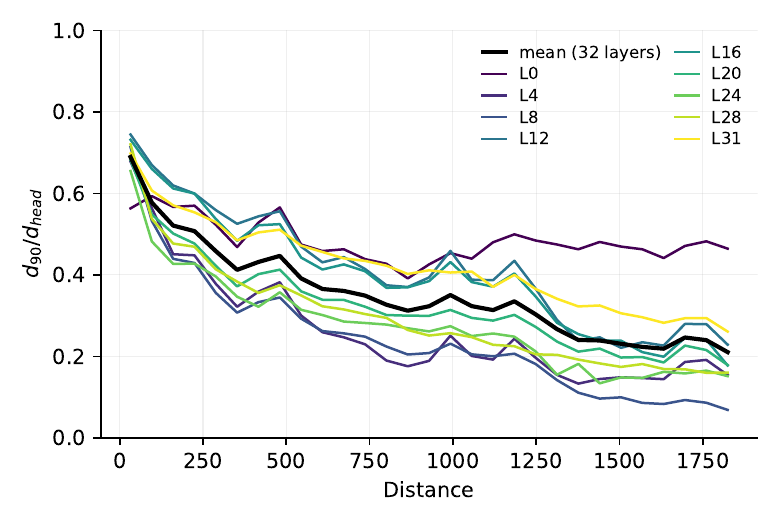}
    \caption{Query-conditioned effective dimensionality across token distance for the pretrained Pythia-2.8B-deduped checkpoint. For clarity, we show nine representative layers from L0 to L31, together with the mean across all 32 layers. Each curve reports $d_{90}/d_{\text{head}}$ averaged across attention heads, where $d_{90}$ is the number of principal components required to explain 90\% of the variance in the attention-weighted value aggregates. Relative token distances range from 1 to 1,856.}    
    \label{fig:effective-dim}
\end{figure}

As an empirical check of this intuition, we examine whether the information used by attention exhibits a distance-dependent pattern in pretrained models. For each query, we aggregate attention-weighted value contributions over distance bins and measure the number of principal components required to explain 90\% of the variance. As shown in Figure~\ref{fig:effective-dim}, the effective dimensionality generally decreases with token distance, indicating that the variation in information from distant context is concentrated in fewer directions. This further motivates our hypothesis that local and distant context require different representational capacities and distant context may require lower representational dimensionality than local context. If this hypothesis holds, can we reduce the dimensionality of KV representations as token distance increases without substantially harming model performance?

While prior studies have extensively explored the KV cache reduction problem, none of them has addressed the aforementioned question directly. We categorize the relevant literature into two distinct categories. The first maintains a local context window while sparsifying attention over distant tokens. Specifically, KV cache eviction methods, e.g., sliding-window attention~\cite{beltagy2020longformer}, StreamingLLM~\cite{xiao2023efficient}, and H2O~\cite{zhang2023h2o}—systematically discard past tokens based on varying importance criteria. All of them, however, retain a span of recent tokens that are guaranteed not to be evicted, suggesting that information carried by local tokens is relatively more important for prediction. The second approach modifies the model architecture itself to reduce representational dimensionality. Multi-head Latent Attention (MLA)~\cite{liu2024deepseek}, proposed by DeepSeek, applies uniform low-rank compression across all past tokens, allowing the model to adapt to this low-rank regime through pretraining.  Although MLA reduces memory overhead, its uniform latent dimensionality treats local and distant tokens identically. Compressed Sparse Attention (CSA) in DeepSeek-V4 ~\cite{deepseek2026v4} reduces KV cache further by compressing multiple tokens horizontally into one token.   Taken together, prior work has yet to characterize how token distance influences the dimensionality required for attention. This motivates our investigation of the hypothesis that \textit{representational capacity should be allocated based on token distance rather than applied uniformly}. We refer to this principle as Distance-Adaptive Representation (DAR)\footnote{Project page and code:
\url{https://zywang624.github.io/distance-adaptive-representation/}}.

\begin{figure}[t]
    \centering
    \includegraphics[
        width=0.8\linewidth,
        trim={105 90 115 95},
        clip
    ]{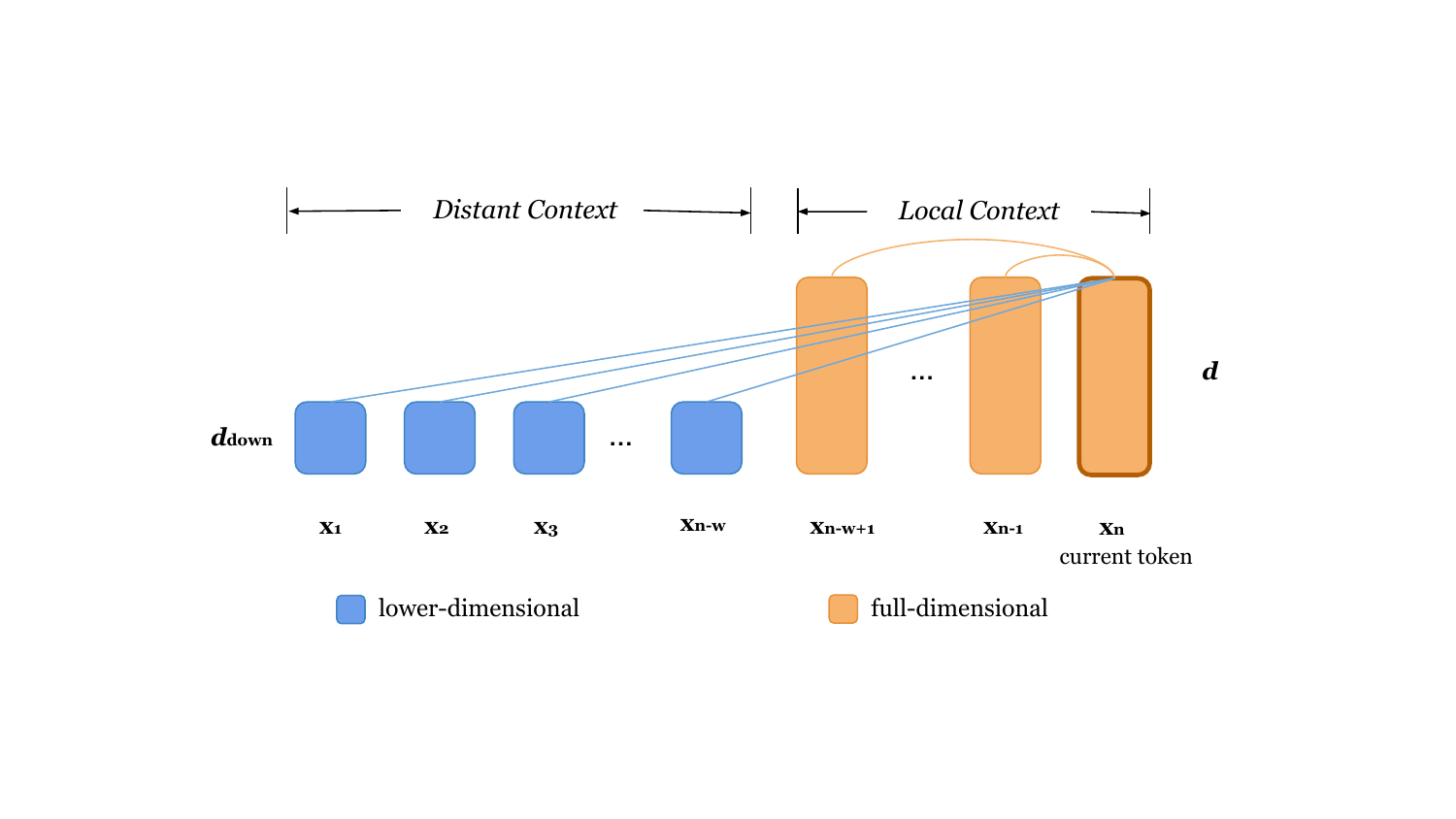}
    \caption{Tokens within a local window of size $w$ (including the current token $x_n$) are represented at dimensionality $d$, while tokens beyond the window are represented at a lower dimensionality $d_{\text{down}}$. The current token attends to all preceding tokens.}
    \label{fig:hypothesis}
\end{figure}

To verify this hypothesis, we adopt an implementation of DAR that partitions the preceding sequence into a \textbf{local context} within a recent window and a \textbf{distant context} beyond the window, assigning full-dimensional representations to the former and lower-dimensional representations to the latter, as illustrated in Figure~\ref{fig:hypothesis}. Our main findings are as follows:

\begin{itemize}
    \item The dimensionality assigned to distant tokens can be reduced without substantially harming perplexity performance, while uniform reduction leads to larger degradation. This pattern holds across 70M, 160M, and 410M pretraining scales, where distance-adaptive representation remains comparable to the full-dimensional baseline and consistently outperforms uniform reduction.
    \item The hypothesis extends beyond pretraining perplexity: when applied as continued supervised fine-tuning on a 1B-scale model, distance-adaptive dimensionality preserves downstream task performance.
    \item We develop DAR-opt, an MLA-style efficient realization of DAR that preserves model quality while reducing the average KV-cache size by 68.8\% at 8K context and achieving up to a $1.9\times$ decoding speedup at 128K on the evaluated models.
\end{itemize}

\section{Distance-Adaptive Representation}

Distance-Adaptive Representation (DAR) is a principle that allocates representational capacity in attention according to token distance from the prediction target. We use a two-regime partition to evaluate this hypothesis. For each query, we define the most recent $w$ positions as the local context and tokens beyond this window as the distant context. Under this scheme, full dimensionality is assigned to the local context, while a lower dimensionality is used for the distant context.

\subsection{Bottleneck Representation for Distant Tokens}

For each token at position $j$, let $\mathbf{h}_j \in \mathbb{R}^{d}$ denote its hidden state. To realize the two-regime partition, we keep the original hidden state $\mathbf{h}_j$ for tokens within a local window of $w$, and produce a lower-dimensional alternative for distant tokens through a linear projection:
\begin{equation}
\label{eq:down_proj}
    \mathbf{h}_{j}^{D} = \mathbf{h}_j \mathbf{W}_{\text{down}},
\end{equation} 
where $\mathbf{h}_{j}^{D} \in \mathbb{R}^{d_{\text{down}}}$, $\mathbf{W}_{\text{down}} \in \mathbb{R}^{d \times d_{\text{down}}}$, and $d_{\text{down}} < d$. The bottleneck dimensionality $d_{\text{down}}$ controls the representational capacity assigned to distant context. We use $\mathbf{h}_{j}^{D}$ as the underlying representation for distant tokens when accessed in attention. Similar to MLA~\cite{liu2024deepseek}, $\mathbf{h}_{j}^{D}$ can be viewed as a compressed latent representation, while the key difference is that tokens within the local window retain full dimensionality.

\nop{, although the same principle can also be applied when local tokens use a higher-dimensional compressed representation.}
\nop{This treatment is consistent with MLA \cite{liu2024deepseek}; $\mathbf{h}_{j}^{D}$ can be interpreted as a compressed latent vector. The key difference is that tokens within the local window retain full dimensionality (though, in principle, they could also use a  compressed representation).}

\subsection{Hybrid Attention over Two Representations}
\label{sec:hybird_attn}

Given a query $\mathbf{q}_i$ at position $i$, preceding tokens use different representations depending on their distance from the query, namely $\mathbf{h}_j$ for local tokens and $\mathbf{h}_j^D$ for distant tokens. Attention therefore proceeds along two paths: a local path for tokens within the local window and a global path for distant tokens. To share the same key and value projections $\mathbf{W}_K$ and $\mathbf{W}_V$ across both paths, we project the bottlenecked representation $\mathbf{h}_j^D$ back to the model dimension $d$ before computing keys and values along the global path. For clarity, we present the formulation with a single attention head and omit standard operations such as layer normalization; the multi-head formulation follows directly.

For tokens beyond the window, the bottlenecked representation 
$\mathbf{h}_{j}^{D}$ is first projected back to dimension $d$:
\begin{equation}
\label{eq:up_proj}
    \mathbf{h}_{j}' = \mathbf{h}_{j}^{D} \, \mathbf{W}_{\text{up}},
\end{equation}
where $\mathbf{W}_{\text{up}} \in \mathbb{R}^{d_{\text{down}} \times d}$. This up-projection maps the compressed representation back to the model dimension without recovering information removed by the bottleneck, allowing the local and global paths to share the same key and value projections.

For each preceding position $j$, the keys and values used in attention 
are then computed based on its distance from the query position $i$:
\begin{equation}
\label{eq:k_v_select}
    \mathbf{k}_j =
    \begin{cases}
        \operatorname{RoPE}(\mathbf{h}_j \mathbf{W}_K), & \text{if } i - j < w, \\
        \operatorname{RoPE}(\mathbf{h}_j' \mathbf{W}_K), & \text{otherwise,}
    \end{cases}
    \quad
    \mathbf{v}_j =
    \begin{cases}
        \mathbf{h}_j \mathbf{W}_V, & \text{if } i - j < w, \\
        \mathbf{h}_j' \mathbf{W}_V, & \text{otherwise,}
    \end{cases}
\end{equation}
where $\operatorname{RoPE}(\cdot)$ applies rotary position embeddings and $w$ is the size of the local window. The attention output for query $\mathbf{q}_i$ is then computed in the standard way:
\begin{equation}
    \mathbf{o}_i = \operatorname{Softmax}\!\left(
        \frac{\mathbf{q}_i \mathbf{K}_i^\top}{\sqrt{d_k}}
    \right)\mathbf{V}_i,
\end{equation}
where $\mathbf{K}_i = [\mathbf{k}_1; \dots; \mathbf{k}_i]$, $\mathbf{V}_i = [\mathbf{v}_1; \dots; \mathbf{v}_i]$, and $d_k$ is the per-head key dimensionality of the underlying multi-head attention. As in standard attention, the query is computed as $\mathbf{q}_i = \operatorname{RoPE}(\mathbf{h}_i \mathbf{W}_Q)$, and the attention output $\mathbf{o}_i$ is further projected by an output projection $\mathbf{W}_O$ before being passed to the next layer. The window size $w$ thus serves as the boundary between the two paths, determining whether a preceding token is attended to via the original representation or via the bottlenecked representation.

\subsection{Training and Inference}
During training, each past token maintains both its original and bottlenecked representations. Each query attends to all past tokens, with the appropriate key and value representations selected based on distance  (Eq.~\eqref{eq:k_v_select}). The standard next-token prediction objective is used:
\begin{equation}
    \mathcal{L} = -\sum_{t=1}^{T} \log P(x_t \mid x_{<t};\, \theta),
\end{equation}
where $x_t$ is the $t$-th token, $x_{<t}$ denotes all preceding  tokens, $T$ is the sequence length, and $\theta$ denotes all model  parameters. No auxiliary losses or additional supervision signals  are introduced; the bottleneck projections $\mathbf{W}_{\text{down}}$ and $\mathbf{W}_{\text{up}}$ are updated through positions that use the global path.

During inference, tokens within the local window retain both local- and global-path key and value states. Once a token moves outside the window, its local-path states can be discarded, since subsequent queries access it only through the global path.

\subsection{Efficient Inference with DAR-opt}
\label{sec:efficient-inference}

The reference implementation above isolates the effect of distance-dependent representation dimensionality for hypothesis validation. We further develop \textbf{DAR-opt}, an efficient realization that exploits the efficiency potential of DAR. DAR-opt adopts decoupled RoPE~\cite{liu2024deepseek}, separating attention into content and rotary components while varying the content dimensionality according to token distance.

Following MLA~\cite{liu2024deepseek}, for each token we construct a local content latent $\mathbf{c}_j^L \in \mathbb{R}^{d_c^L}$ and a distant content latent $\mathbf{c}_j^D \in \mathbb{R}^{d_c^D}$, with $d > d_c^L > d_c^D$. Both are projected from the hidden state $\mathbf{h}_j$ and are used to compute the content key and value components. Positional information is represented separately by a rotary query $\mathbf{q}_i^R \in \mathbb{R}^{d_r}$ and a rotary key $\mathbf{k}_j^R \in \mathbb{R}^{d_r}$, where $d_r$ denotes the dimensionality of the rotary component. Rather than caching full-dimensional keys and values, DAR-opt caches these compressed latent representations, with a higher-dimensional latent for local tokens and a lower-dimensional latent for distant tokens.

During decoding, tokens within the local window cache both $\mathbf{c}_j^L$ and $\mathbf{c}_j^D$, together with the rotary key $\mathbf{k}_j^R$. Once a token moves outside the window, $\mathbf{c}_j^L$ can be discarded, while $\mathbf{c}_j^D$ and $\mathbf{k}_j^R$ are retained for the distant context. For a sequence of length $T$, the average cache size per token is therefore
\begin{equation}
C_{\mathrm{DAR\text{-}opt}}
=
d_c^D + d_r + \frac{d_c^L w}{T}.
\end{equation}
As the context length increases, the local-window overhead is amortized and the average cache size approaches $d_c^D+d_r$. Detailed formulations of the attention computation and weight absorption are provided in Appendix~\ref{app:dar-opt}.

\section{Experiments}

We evaluate DAR through a series of analyses and experiments. We first examine whether attention-weighted value information exhibits a distance-dependent effective dimensionality pattern, and then directly validate the two-regime hypothesis at the Pythia-70M scale by comparing distance-adaptive and uniform dimensionality reduction. We further examine whether this behavior holds across Pythia-70M, 160M, and 410M models. Next, we apply DAR to continued supervised fine-tuning of OLMo-2-1B-SFT to evaluate its effect on downstream task performance. Finally, we evaluate DAR-opt, an MLA-style efficient realization of DAR, through from-scratch pretraining and long-context inference measurements. We additionally study its application to a pretrained MLA-based LLaMA-3-8B model in Appendix~\ref{app:dar-mla}.

\nop{
We conduct pretraining and supervised fine-tuning experiments to validate the two-regime partition scheme described above. If DAR is effective, the two-regime partition scheme should perform close to full-dimensional attention, and substantially better than uniform lower-dimensional attention applied to all tokens.} 

\paragraph{Hypothesis validation and scaling.}
For the controlled hypothesis validation and scaling experiments, we pretrain models from scratch following the Pythia training recipe~\cite{gpt-neox-library,biderman2023pythia}, with a maximum sequence length of 2,048 tokens. The hypothesis validation experiments use the Pythia-70M architecture, while the scaling analysis additionally includes Pythia-160M and Pythia-410M. All models are trained for 19,073 steps with a batch size of 256 on the first 10B tokens of the deduplicated Pile~\cite{biderman2022datasheet,gao2020pile}, above the compute-optimal token count for models at these scales~\cite{hoffmann2022training}. Unless otherwise specified, DAR uses a local window size of $w=128$. Performance is evaluated by perplexity on subsets of FineWeb-Edu~\cite{lozhkov2024fineweb-edu}, WikiText-103~\cite{merity2016pointer}, and C4~\cite{raffel2020exploring}. Most evaluation sequences extend well beyond the local window size $w$, substantially exercising the distant-context path. Evaluation details, including document-length distributions and rolling evaluation for long documents, are provided in Appendix~\ref{app:evaluation}. We additionally study the effect of the local window size in Appendix~\ref{app:window-ablation}.

\paragraph{Downstream task evaluation.}
To evaluate whether the findings extend beyond pretraining perplexity, we start from OLMo-2-1B-SFT~\cite{olmo20242olmo2furious} and perform continued supervised fine-tuning with DAR. Training proceeds in two stages, each consisting of one epoch over the T\"ulu 3 dataset used for OLMo-2-1B-SFT~\cite{lambert2024tulu}. In the first stage, only the bottleneck parameters $\{\mathbf{W}_{\text{down}}, \mathbf{W}_{\text{up}}\}$ are trained while the remaining parameters are frozen. In the second stage, all parameters are trained jointly. We use AdamW with a linear learning-rate schedule, a warmup ratio of $0.03$, a batch size of $512$, and a maximum sequence length of $2{,}048$. The first and second stages use learning rates of $3\times10^{-4}$ and $3\times10^{-5}$, respectively. Performance is evaluated using \texttt{lm-evaluation-harness}~\cite{eval-harness} on six downstream benchmarks described in Section~\ref{sec:downstream}.

\paragraph{Efficient realization.}
We evaluate DAR-opt through from-scratch pretraining at the Pythia-70M scale and long-context inference measurements. We compare DAR-opt with uniform MLA baselines in terms of model quality, cache usage, and decoding latency. An additional experiment applying distance-adaptive latent allocation to a pretrained MLA-based LLaMA-3-8B model is provided in Appendix~\ref{app:dar-mla}.

\subsection{Core Hypothesis Validation}
\label{sec:hypothesis_validation}

We validate the core hypothesis from two complementary perspectives. First, we examine whether the information used by attention exhibits a distance-dependent dimensionality pattern. Second, we test whether explicitly allocating representational capacity by distance performs better than applying the same dimensionality reduction uniformly across all token positions.

\paragraph{Analysis across token distance.}
We perform a query-conditioned analysis on official pretrained models. For each query position $i$ and distance bin $b$, we aggregate the value vectors within the bin using their attention weights:
\begin{equation}
    \mathbf{m}_i^{(b)}
    =
    \sum_{j:\, i-j\in b} a_{ij}\mathbf{v}_j.
\end{equation}
The resulting vector describes the information contributed to query $i$ from tokens at a particular distance range. We collect these vectors across queries and measure how many directions are needed to explain their variation. Specifically, for each layer and attention head, we apply PCA and report $d_{90}$, the smallest number of principal components required to explain 90\% of the variance, normalized by the head dimension $d_{\text{head}}$. We average the normalized value across all attention heads within each layer. As shown in Figure~\ref{fig:effective-dim}, this quantity generally decreases with token distance, indicating that attention-weighted value information from more distant context varies along fewer effective directions. We observe consistent trends across other pretrained models. Full analysis settings are provided in Appendix~\ref{app:effective-dim}.

\paragraph{DAR versus uniform reduction.}
We use the Pythia-70M architecture and vary the bottleneck dimension $d_{\text{down}}$ under a fixed window size $w=128$, and compare against two reference points: (i) a full-dimensional baseline ($d=512$, ``Vanilla''), and (ii) a uniform reduction baseline that applies the same lower dimensionality $d_{\text{down}}$ to all tokens regardless of distance.

Table~\ref{tab:hypothesis} reports perplexity across the three evaluation corpora. Two observations support the hypothesis. First, DAR with $d_{\text{down}}=256$ and $d_{\text{down}}=128$ achieves perplexity comparable to the full-dimensional baseline; only at $d_{\text{down}}=64$ does noticeable degradation appear (Rel.\ 101.57\%). This suggests that distant-token representations can tolerate substantial dimensionality reduction under the evaluated setting. Second, when the lower-dimensional representations are applied uniformly across all token positions, performance degrades more sharply than with DAR. The gap between DAR and uniform reduction shows that reducing dimensionality only for distant tokens is substantially less harmful than applying the same reduction uniformly across all token positions.
For the $d_{\text{down}}=128$ setting, we additionally run pretraining with three random seeds. Relative to the mean Vanilla perplexity across seeds, Vanilla, DAR, and Uniform Reduction obtain $100.00 \pm 0.73\%$, $99.68 \pm 0.57\%$, and $103.82 \pm 0.91\%$, respectively. The perplexities along the training trajectories are provided in Appendix~\ref{app:training-curves}.

\nop{
Figure~\ref{fig:ppl_rel} shows the relative perplexity trajectory across pretraining. In the early stages, all variants exhibit elevated perplexity relative to Vanilla, but the gap closes at different rates. DAR with $d_{\mathrm{down}}\in\{128, 256\}$ converges to Vanilla by the end of training, while DAR with $d_{\mathrm{down}}=64$ remains slightly above. Uniform reduction remains above Vanilla throughout training across all $d_{\mathrm{down}}$ values, with the gap widening as $d_{\mathrm{down}}$ decreases. At each $d_{\mathrm{down}}$, DAR outperforms Uniform Reduction throughout pretraining.\nop{demonstrating that the dimensional asymmetry holds through the training trajectory.}
}

\begin{table}[ht]
\centering
\small
\setlength{\tabcolsep}{6pt}
\caption{DAR validation at the Pythia-70M scale. DAR is run with window size $w=128$ across all bottleneck dimensions. Perplexity is reported on three evaluation corpora: a subset of FineWeb-Edu, C4 and WikiText-103. Rel. is the average per-dataset perplexity ratio relative to Vanilla, reported as a percentage (lower is better).} 
\label{tab:hypothesis}
\begin{tabular}{llccccc}
\toprule
\textbf{Model} & \textbf{Config} & \textbf{FineWeb-Edu} & \textbf{C4} & \textbf{WikiText-103} & \textbf{Rel.(\%)} \\
\midrule
Vanilla & $d{=}512$ & 119.58 & 145.52 & 52.58 & 100.00 \\
\midrule
\multirow{3}{*}{DAR}
& $d_{\mathrm{down}}{=}256$ & 118.26 & 144.14 & 52.03 & 98.97 \\
& $d_{\mathrm{down}}{=}128$ & 118.27 & 143.99 & 52.32 & 99.12 \\
& $d_{\mathrm{down}}{=}64$  & 121.21 & 146.97 & 53.81 & 101.57 \\
\midrule
\multirow{3}{*}{Uniform reduction}
& $d_{\mathrm{down}}{=}256$ & 122.74 & 148.24 & 53.97 & 102.39 \\
& $d_{\mathrm{down}}{=}128$ & 124.06 & 149.91 & 54.57 & 103.52 \\
& $d_{\mathrm{down}}{=}64$ & 133.38 & 160.02 & 57.92 & 110.56 \\
\bottomrule
\end{tabular}
\end{table}

\nop{
\begin{table}[t]
\centering
\small
\setlength{\tabcolsep}{6pt}
\caption{DAR validation at the Pythia-70M scale. DAR is run with window size $w=128$ across all bottleneck dimensions. Perplexity is reported on three evaluation corpora: a subset of FineWeb-Edu, C4 and WikiText-103. Rel. is the average per-dataset perplexity ratio relative to Vanilla, reported as a percentage (lower is better).} 
\label{tab:hypothesis}
\begin{tabular}{llccccc}
\toprule
\textbf{Model} & \textbf{Config} &  \textbf{FineWeb-Edu} & \textbf{C4} & \textbf{WikiText-103} & \textbf{Rel.(\%)} \\
\midrule
Vanilla & $d{=}512$ & 119.65 & 145.53 & 52.91 & 100.00 \\
\midrule
\multirow{3}{*}{DAR}
& $d_{\mathrm{down}}{=}256$ & 118.27 & 144.10 & 52.26 & 98.90 \\
& $d_{\mathrm{down}}{=}128$ & 119.62 & 145.50 & 52.92 & 100.01 \\
& $d_{\mathrm{down}}{=}64$  & 121.92 & 148.27 & 54.27 & 102.14 \\
\midrule
\multirow{3}{*}{Uniform reduction}
& $d_{\mathrm{down}}{=}256$ & 120.66 & 145.98 & 53.10 & 100.53 \\
& $d_{\mathrm{down}}{=}128$ & 126.02 & 151.80 & 55.15 & 104.65 \\
& $d_{\mathrm{down}}{=}64$ & 134.19 & 162.94 & 58.32 & 111.47 \\
\bottomrule
\end{tabular}
\end{table}
}

\subsection{Generalization Across Pretraining Scales}

To examine whether the same observation holds at larger pretraining scales, we extend the experiment to Pythia-160M and Pythia-410M. We compare DAR with both the corresponding full-dimensional baseline and Uniform Reduction at each scale. The bottleneck dimension is fixed at $d_{\text{down}}=d/4$ across all scales.

As shown in Table~\ref{tab:scaling_analysis}, DAR remains comparable to the full-dimensional baseline and consistently achieves lower perplexity than Uniform Reduction under the same bottleneck ratio. This indicates that the dimensional asymmetry between local and distant tokens is not limited to the 70M setting and persists across the evaluated model scales. Within the evaluated scale range, DAR can preserve competitive performance using the same relative ratio, $d_{down}=d/4$, providing further evidence that distant context does not require full-dimensional representation. \nop{distant-token representations may not require full dimensionality, although larger-scale experiments are needed to determine how this trend holds more generally.}

\begin{table}[t]
\centering
\small
\setlength{\tabcolsep}{6pt}
\caption{Generalization of DAR across pretraining scales. DAR uses
$w=128$ and $d_{\text{down}} = d/4$ at each scale. Perplexity is
reported on a subset of FineWeb-Edu, C4 and WikiText-103. Rel.(\%) is the average per-dataset perplexity ratio relative
to the Vanilla model at the same scale (lower is better).}
\label{tab:scaling_analysis}
\begin{tabular}{llccccc c}
\toprule
\textbf{Model} & \textbf{Config} & \textbf{FineWeb-Edu} & \textbf{C4} & \textbf{WikiText-103} & \textbf{Rel.(\%)} \\
\midrule
\multirow{2}{*}{Pythia-70M}
& Vanilla ($d{=}512$) & 119.58 & 145.52 & 52.58 & 100.00 \\
& DAR ($d_{\mathrm{down}}{=}128$) & 118.27 & 143.99 & 52.32 & 99.12 \\
& Uniform ($d_{\mathrm{down}}{=}128$) & 124.06 & 149.91 & 54.57 & 103.52 \\
\midrule
\multirow{2}{*}{Pythia-160M}
& Vanilla ($d{=}768$) & 79.46 & 99.51 & 35.96 & 100.00 \\
& DAR ($d_{\mathrm{down}}{=}192$) & 78.93 & 97.84 & 36.05 & 99.30 \\
& Uniform ($d_{\mathrm{down}}{=}192$) & 81.81 & 100.66 & 36.92 & 102.26 \\
\midrule
\multirow{2}{*}{Pythia-410M}
& Vanilla ($d{=}1024$) & 61.71 & 76.61 & 28.94 & 100.00 \\
& DAR ($d_{\mathrm{down}}{=}256$) & 62.23 & 76.90 & 29.31 & 100.83 \\
& Uniform ($d_{\mathrm{down}}{=}256$) & 63.65 & 78.96 & 29.52 & 102.74 \\
\bottomrule
\end{tabular}
\end{table}

\subsection{Effect on Downstream Tasks}
\label{sec:downstream}

To further examine whether DAR preserves task-level performance, we evaluate it on a suite of downstream benchmarks under different bottleneck dimensions $d_{\text{down}}$, while keeping the window size fixed at $w=128$. To isolate the effect of $d_{\text{down}}$ from the effect of introducing the bottleneck module itself, we use the same DAR architecture across all configurations and treat the setting $d_{\text{down}} = d = 2048$ as the no-bottleneck baseline; this configuration includes the same down-projection and up-projection modules as the other configurations, but applies no actual dimensionality reduction. 

We evaluate on MMLU~\cite{hendryckstest2021} for massive multitask 
understanding, HellaSwag~\cite{zellers2019hellaswag} for commonsense inference, CommonsenseQA~\cite{talmor-etal-2019-commonsenseqa} for commonsense question answering, GSM8K~\cite{cobbe2021training} for mathematical reasoning, MBPP~\cite{austin2021program} for code generation, and Multi-News from LongBench~\cite{bai2024longbench} for multi-document summarization. We employ a 5-shot setting for MMLU, HellaSwag, CommonsenseQA, and GSM8K, a 3-shot setting for MBPP, and a zero-shot setting for Multi-News. The reported metrics are accuracy (Acc) on MMLU and CommonsenseQA, normalized accuracy (Acc-norm) on HellaSwag, flexible-extract match on GSM8K, Pass@1 on MBPP, and ROUGE scores on Multi-News. To ensure the global path is engaged during evaluation, we exclude samples whose input context is shorter than the window size. The average input context lengths for the six tasks are $742$, $532$, $332$, $939$, $673$, and $1{,}394$ tokens, respectively.

\begin{table}[t]
\centering
\small
\setlength{\tabcolsep}{3pt}
\caption{Downstream task evaluation. All configurations use the DAR architecture with $w=128$. The first row, with $d_{\text{down}} = d = 2048$, applies no actual dimensionality reduction and serves as the no-bottleneck baseline; subsequent rows progressively reduce $d_{\text{down}}$. Avg. is the average across the six benchmarks. Rel.(\%) is the average of task-specific relative scores compared to the no-bottleneck baseline (higher is better).}
\label{tab:downstream}
\begin{tabular}{lcccccccc}
\toprule
\textbf{Config} & \textbf{MMLU} & \textbf{HellaSwag} & \textbf{CSQA}
& \textbf{GSM8K} & \textbf{MBPP} & \textbf{Multi-News} & \textbf{Avg.} & \textbf{Rel.(\%)} \\
\midrule
\nop{
$d_{\mathrm{down}}{=}2048$ & 0.4131 & 0.6500 & 0.5315 & 0.4276 & 0.1500 & 0.1800 & 0.3921 & 100.00 \\
\midrule
$d_{\mathrm{down}}{=}1024$ & 0.3994 & 0.6522 & 0.5152 & 0.4450 & 0.1480 & 0.1863 & 0.3910 & 100.02 \\
$d_{\mathrm{down}}{=}512$  & 0.3968 & 0.6469 & 0.5291 & 0.4253 & 0.1260 & 0.1987 & 0.3871 & 98.16 \\
$d_{\mathrm{down}}{=}256$  & 0.3957 & 0.6418 & 0.5283 & 0.4109 & 0.0900 & 0.1663 & 0.3722 & 90.39 \\
$d_{\mathrm{down}}{=}128$  & 0.4069 & 0.6390 & 0.5373 & 0.4056 & 0.0800 & 0.1539 & 0.3705 & 88.59 \\
$d_{\mathrm{down}}{=}64$   & 0.3866 & 0.6329 & 0.4922 & 0.3912 & 0.0420 & 0.0913 & 0.3394 & 75.62 \\
}
$d_{\mathrm{down}}{=}2048$ & 0.4044 & 0.6502 & 0.5315 & 0.4450 & 0.1320 & 0.1653 & 0.3881 & 100.00 \\
\midrule
$d_{\mathrm{down}}{=}1024$ & 0.3989 & 0.6493 & 0.5214 & 0.4496 & 0.1420 & 0.1675 & 0.3881 & 101.09 \\
$d_{\mathrm{down}}{=}512$ & 0.4051 & 0.6462 & 0.5304 & 0.4114 & 0.1440 & 0.1855 & 0.3871 & 102.19 \\
$d_{\mathrm{down}}{=}256$ & 0.3978 & 0.6440 & 0.5228 & 0.4182 & 0.1180 & 0.1867 & 0.3812 & 98.68 \\
$d_{\mathrm{down}}{=}128$ & 0.4067 & 0.6375 & 0.5209 & 0.4008 & 0.0780 & 0.1388 & 0.3638 & 88.29 \\
$d_{\mathrm{down}}{=}64$ & 0.3891 & 0.6338 & 0.5012 & 0.3907 & 0.0300 & 0.1545 & 0.3499 & 82.00 \\
\bottomrule
\end{tabular}
\vspace{2pt}
\begin{flushleft}
\end{flushleft}
\end{table}


Table~\ref{tab:downstream} reports the per-task scores. Two observations stand out. First, DAR preserves downstream performance under moderate dimensionality reduction. At $d_{\text{down}}=1024$ ($d/2$) and $d_{\text{down}}=512$ ($d/4$), the relative scores are 101.09\% and 102.19\%, respectively. This is consistent with the perplexity results in Section~\ref{sec:hypothesis_validation}, where DAR also matched the full-dimensional baseline when $d_{\text{down}}=d/4$. This suggests that, for the evaluated tasks and scale, distant-token representations can tolerate substantial dimensionality reduction. Second, downstream tasks become more sensitive under more aggressive compression. At $d_{\text{down}}=256$ ($d/8$), DAR still retains 98.68\% of the baseline score, while the relative score drops to 88.29\% at $d_{\text{down}}=128$ ($d/16$). This suggests that some tasks become more vulnerable once the distant-token representation is compressed too aggressively.


\subsection{Efficient Realization of DAR}
\label{sec:efficient-eval}

Building on the distance-adaptive representation principle, we develop DAR-opt for efficient inference using decoupled RoPE and weight absorption. We evaluate its model quality, cache usage, and decoding efficiency at the Pythia-70M scale. We further validate the distance-adaptive allocation on an MLA-converted LLaMA-3-8B model through distillation, where DAR-opt retains 97.29\% of the teacher performance; details are provided Appendix~\ref{app:dar-mla}.

\paragraph{Model quality.}
We pretrain DAR-opt from scratch at the Pythia-70M scale to verify that the distance-adaptive principle extends to the MLA-style efficient formulation. MLA uses a content latent of uniform dimension $d_c$. In DAR-opt, tokens within the local window use a higher-dimensional content latent with $d_c^L=480$, while tokens outside the window use a lower-dimensional content latent with $d_c^D=120$, under a local window of $w=128$. During caching, local tokens additionally retain the local latent $c^L$, whereas distant tokens retain only $c^D$. A rotary component of dimension $d_r=32$ is used for all models in this experiment. We compare DAR-opt against uniform MLA baselines with different content latent dimensions.

For a context length $T$, the average cached dimensionality per token is
\[
C_{\mathrm{MLA}} = d_c + d_r,
\qquad
C_{\mathrm{DAR\text{-}opt}}
= d_c^D + d_r + \frac{d_c^L w}{T}.
\]
At $T=2048$, DAR-opt therefore requires $182$ dimensions per token, exactly matching MLA-150 and providing a matched-cache comparison.

As shown in Table~\ref{tab:dar-opt-pretrain}, DAR-opt achieves model quality comparable to the higher-capacity MLA-480 baseline while using a substantially smaller average cache. Under the same average cache budget of 182 dimensions per token, DAR-opt substantially outperforms MLA-150 (Rel.\ 98.16\% vs.\ 101.40\%), demonstrating a better performance--cache trade-off than uniform latent compression.

\begin{table}[t]
    \centering
    \small
    \setlength{\tabcolsep}{4pt}
    \caption{From-scratch evaluation of DAR-opt at the Pythia-70M scale. MLA baselines use the same content latent dimension for local and distant tokens, while DAR-opt uses $d_c^L=480$ for local tokens and $d_c^D=120$ for distant tokens. MLA-480 serves as the full-capacity MLA baseline, while MLA-150 provides a matched-cache baseline with the same average cache size as DAR-opt. Avg.\ cache size is measured per token at a context length of 2,048. Rel.\ is the average per-dataset perplexity ratio relative to MLA-480 (lower is better).}
    \label{tab:dar-opt-pretrain}
    \begin{tabular}{lccc|c|cccc}
        \toprule
        \textbf{Method}
        & \multicolumn{3}{c|}{\textbf{Latent configuration}}
        & \textbf{Avg.\ cache size}
        & \multicolumn{4}{c}{\textbf{Perplexity}} \\
        \cmidrule(lr){2-4}
        \cmidrule(lr){6-9}
        &
        \textbf{$d_c^L$}
        & \textbf{$d_c^D$}
        & \textbf{$d_r$}
        &
        & \textbf{FineWeb-Edu}
        & \textbf{C4}
        & \textbf{WikiText-103}
        & \textbf{Rel.(\%)} \\
        \midrule
        MLA-480
        & 480 & 480 & 32
        & 512
        & 118.37 & 143.93 & 51.93 & 100.00 \\
        \nop{
        MLA-192
        & 192 & 192 & 32
        & 224
        & 120.40 & 145.92 & 53.25 & XXX \\
        }
        MLA-150
        & 150 & 150 & 32
        & 182
        & 120.29 & 144.49 & 53.06 & 101.40 \\
        MLA-120
        & 120 & 120 & 32
        & 152
        & 123.65 & 148.49 & 54.25 & 104.03 \\
        \midrule
        \textbf{DAR-opt}
        & \textbf{480} & \textbf{120} & \textbf{32}
        & \textbf{182}
        & \textbf{115.92} & \textbf{141.70} & \textbf{50.95} & \textbf{98.16} \\
        \bottomrule
    \end{tabular}
\end{table}

\begin{figure}[t]
    \centering
    \begin{subfigure}[t]{0.49\linewidth}
        \centering
        \includegraphics[width=\linewidth]{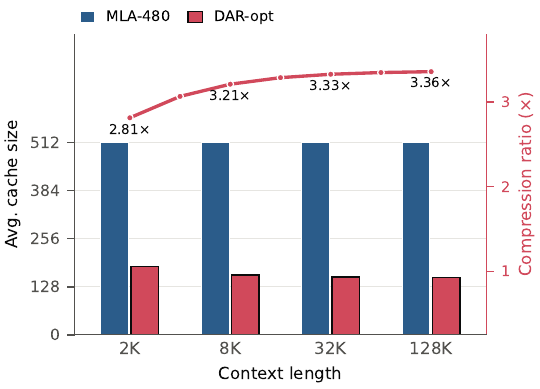}
        \caption{Average cache size and compression ratio.}
    \end{subfigure}
    \hfill
    \begin{subfigure}[t]{0.49\linewidth}
        \centering
        \includegraphics[width=\linewidth]{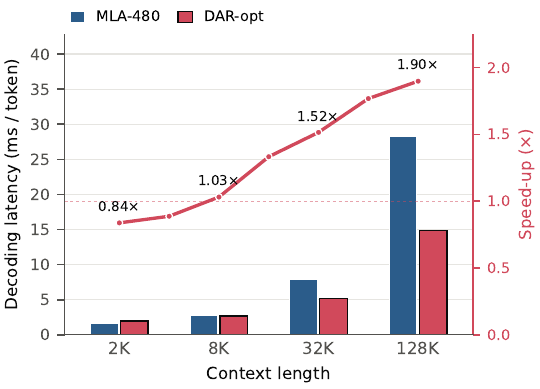}
        \caption{Decoding latency and speedup.}
    \end{subfigure}
    \caption{Inference efficiency of DAR-opt across context lengths. Compression ratios and speedups are measured relative to MLA-480. DAR-opt uses $d_c^L=480$, $d_c^D=120$, $d_r=32$, and $w=128$.}
    \label{fig:dar-opt-efficiency}
\end{figure}

\paragraph{Inference efficiency.}
We further evaluate the cache usage and decoding efficiency of DAR-opt at the Pythia-70M scale with longer context lengths. As shown in Figure~\ref{fig:dar-opt-efficiency}, the average cache size of DAR-opt decreases as the context length grows because the local-window overhead is amortized over the sequence. At 8K context, DAR-opt requires 159.5 cached dimensions per token, a $3.21\times$ compression (68.8\% reduction) relative to MLA-480. The compression ratio further approaches $3.36\times$ at 128K context. The reduced cache footprint also translates into faster decoding at long context lengths. Relative to MLA-480, DAR-opt achieves a $1.52\times$ speedup at 32K context and a $1.90\times$ speedup at 128K. We also run the same decoding-efficiency evaluation at Pythia-160M and observe a consistent trend.

\section{Related Work}
Prior studies have extensively explored KV cache reduction through low-rank projection, quantization, key-value sharing, latent attention, compressed sparse attention, sparse attention, and dynamic KV cache eviction. These approaches reduce memory or computation typically by compressing representations or selectively reducing stored or accessed information.

\subsection{Sliding Window and Sparse Attention}
Local window mechanisms have been adopted in many forms. Sliding window attention~\cite{beltagy2020longformer} restricts attention to a window of local tokens, and StreamingLLM~\cite{xiao2023efficient} extends this design with a small set of attention sinks. H2O~\cite{zhang2023h2o} and SnapKV~\cite{snapkv} selectively retain subsets of past tokens based on their estimated importance. SKVQ~\cite{duanmu2024skvq} preserves local tokens at full numerical precision while applying low-bit quantization outside the window. XAttention~\cite{xu2025xattention} and MInference~\cite{jiang2024minference} accelerate long-context inference using sparse attention. Native Sparse Attention (NSA)~\cite{yuan2025native} combines coarse-grained token compression with fine-grained token selection in a trainable sparse-attention architecture, while OpenAI's gpt-oss models~\cite{openai2025gptoss} adopt alternating dense and locally banded sparse attention patterns across layers.

These approaches primarily modify which tokens participate in or are accessed by attention. DAR operates along a different axis: it preserves access to all tokens while varying their representational dimensionality according to distance. These two directions are complementary and could in principle be combined.

\nop{These methods are primarily motivated by reducing the cost of attention or its inference-time footprint. They are training-free and applied at inference time to already-pretrained models.  They do not develop the dual dimensionality proposed in this work. Our work takes a different starting point: we hypothesize that local and distant tokens play structurally distinct roles in next-token prediction, and investigate whether this asymmetry should be reflected in the representational dimensionality. The local window in our setup is a direct design of this hypothesis, not a heuristic inherited from inference-time considerations.}

\subsection{Multi-head Latent Attention}
Recent architectures reduce the size of key and value representations directly during pretraining. Multi-Query Attention (MQA)~\cite{shazeer2019fast} and Grouped-Query Attention (GQA)~\cite{ainslie2023gqa} reduce the number of independent key and value heads, sharing them across queries to lower memory and computation costs. In contrast, Multi-head Latent Attention (MLA)~\cite{liu2024deepseek} compresses per-token representations into a low-rank latent space. Compressed Sparse Attention (CSA)~\cite{deepseek2026v4} further reduces KV cache by compressing hidden states of multiple tokens into one.

Among these methods, MLA is most closely related to our work, as it directly modifies the dimensionality of attention representations.  Our work explores a different aspect of this design space: rather than applying a uniform reduction, we ask whether the required dimensionality should vary with a token’s distance from the prediction target. This perspective suggests an adaptive allocation of representational capacity. \nop{in contrast to the fixed, distance-agnostic schemes adopted by prior approaches.}

\subsection{Multi-Granularity Representation}
Recent advances in representation learning have explored embedding information at multiple levels of granularity within a single vector. Matryoshka Representation Learning (MRL)~\cite{kusupati2022matryoshka} introduces a nested structure that allows a single embedding to be truncated to various sizes while maintaining high accuracy. This concept was then extended to the KV cache in MatryoshkaKV~\cite{lin2024matryoshkakv}, which enables dynamic capacity adjustment during inference through trainable orthogonal projections. FreqKV~\cite{kai2026freqkv} instead compresses KV states in the frequency domain by retaining low-frequency components. These methods exploit different forms of representational structure for compression. In contrast, DAR studies how representational dimensionality should vary with token distance, shifting the focus to an intrinsic structural property of the sequence.

\section{Limitations}
\label{sec:limitation}

Our evaluation remains limited in model scale, architecture coverage, and systems integration. The main pretraining experiments cover decoder-only Transformer models from 70M to 410M parameters, while the downstream study uses a 1B model and the MLA-based extension is evaluated through distillation from a converted LLaMA-3-8B model. The reported DAR-opt systems measurements focus on the 70M scale with context lengths up to 128K using a custom fused Triton kernel, while we additionally observe a consistent decoding-efficiency trend at the 160M scale. The implementation has not yet been integrated into production-oriented serving frameworks such as vLLM. Evaluating DAR across larger models, broader architectures, and more optimized long-context serving settings remains an important direction for future work. Moreover, sequential token distance is natural for autoregressive text but may require redefinition for multimodal or non-sequential inputs.

\section{Conclusion}
We hypothesized that the representational dimensionality required for a token in attention varies with its distance from the prediction target, and introduced Distance-Adaptive Representation (DAR), a principle that allocates representational capacity according to this distance. Through controlled pretraining and supervised fine-tuning experiments, we show that distant tokens can be represented with substantially lower dimensionality without significantly degrading perplexity or downstream task performance, whereas applying the same reduction uniformly across all tokens leads to noticeable performance loss. These results support an asymmetric demand for representational capacity across token distance and challenge the common assumption that attention representations should be uniform across token positions. We further show that this principle can be realized efficiently through DAR-opt and can be incorporated into an MLA-style architecture. We hope this work motivates further investigation into more sophisticated allocations of representational capacity in attention.

\section*{Acknowledgments}
We thank the anonymous reviewers for their insightful suggestions, which helped improve this work. Xuan Luo was partially supported by the BioPACIFIC MIP of the National Science Foundation under Award No. DMR-1933487. We would like to thank Meta for the A100 GPUs used in our experiments. We also gratefully acknowledge the generous support of the NVIDIA Academic Grant Program and NCSA DeltaAI through allocation CIS260864 from the Advanced Cyberinfrastructure Coordination Ecosystem: Services \& Support (ACCESS) program, which is supported by U.S. National Science Foundation.

\bibliographystyle{plain}   
\bibliography{references}   

@inproceedings{
xu2025xattention,
title={{XA}ttention: Block Sparse Attention with Antidiagonal Scoring},
author={Ruyi Xu and Guangxuan Xiao and Haofeng Huang and Junxian Guo and Song Han},
booktitle={Forty-second International Conference on Machine Learning},
year={2025},
url={https://openreview.net/forum?id=KG6aBfGi6e}
}

@inproceedings{
jiang2024minference,
title={{MI}nference 1.0: Accelerating Pre-filling for Long-Context {LLM}s via Dynamic Sparse Attention},
author={Huiqiang Jiang and Yucheng Li and Chengruidong Zhang and Qianhui Wu and Xufang Luo and Surin Ahn and Zhenhua Han and Amir H. Abdi and Dongsheng Li and Chin-Yew Lin and Yuqing Yang and Lili Qiu},
booktitle={The Thirty-eighth Annual Conference on Neural Information Processing Systems},
year={2024},
url={https://openreview.net/forum?id=fPBACAbqSN}
}

@inproceedings{
snapkv,
title={Snap{KV}: {LLM} Knows What You are Looking for Before Generation},
author={Yuhong Li and Yingbing Huang and Bowen Yang and Bharat Venkitesh and Acyr Locatelli and Hanchen Ye and Tianle Cai and Patrick Lewis and Deming Chen},
booktitle={The Thirty-eighth Annual Conference on Neural Information Processing Systems},
year={2024}
}

@misc{deepseek2026v4,
      title={{DeepSeek-V4}: Towards Highly Efficient Million-Token Context Intelligence}, 
      author={DeepSeek-AI},
      year={2026},
      month={April},
      howpublished={Technical Report},
      url={https://huggingface.co/deepseek-ai/DeepSeek-V4-Pro},
}

@inproceedings{biderman2023pythia,
  title={Pythia: A suite for analyzing large language models across training and scaling},
  author={Biderman, Stella and Schoelkopf, Hailey and Anthony, Quentin Gregory and Bradley, Herbie and O’Brien, Kyle and Hallahan, Eric and Khan, Mohammad Aflah and Purohit, Shivanshu and Prashanth, USVSN Sai and Raff, Edward and others},
  booktitle={International Conference on Machine Learning},
  pages={2397--2430},
  year={2023},
  organization={PMLR}
}

@article{gao2020pile,
  title={{The Pile}: An 800gb dataset of diverse text for language modeling},
  author={Gao, Leo and Biderman, Stella and Black, Sid and Golding, Laurence and Hoppe, Travis and Foster, Charles and Phang, Jason and He, Horace and Thite, Anish and Nabeshima, Noa and others},
  journal={arXiv preprint arXiv:2101.00027},
  year={2020}
}

@article{biderman2022datasheet,
  title={Datasheet for the pile},
  author={Biderman, Stella and Bicheno, Kieran and Gao, Leo},
  journal={arXiv preprint arXiv:2201.07311},
  year={2022}
}

@article{hoffmann2022training,
  title={Training compute-optimal large language models},
  author={Hoffmann, Jordan and Borgeaud, Sebastian and Mensch, Arthur and Buchatskaya, Elena and Cai, Trevor and Rutherford, Eliza and Casas, DDL and Hendricks, Lisa Anne and Welbl, Johannes and Clark, Aidan and others},
  journal={arXiv preprint arXiv:2203.15556},
  volume={10},
  year={2022}
}

@software{gpt-neox-library,
  title = {{GPT-NeoX}: Large Scale Autoregressive Language Modeling in PyTorch},
  author = {Andonian, Alex and Anthony, Quentin and Biderman, Stella and Black, Sid and Gali, Preetham and Gao, Leo and Hallahan, Eric and Levy-Kramer, Josh and Leahy, Connor and Nestler, Lucas and Parker, Kip and Pieler, Michael and Phang, Jason and Purohit, Shivanshu and Schoelkopf, Hailey and Stander, Dashiell and Songz, Tri and Tigges, Curt and Thérien, Benjamin and Wang, Phil and Weinbach, Samuel},
  url = {https://www.github.com/eleutherai/gpt-neox},
  doi = {10.5281/zenodo.5879544},
  month = {9},
  year = {2023},
  version = {2.0.0},
}

@misc{eval-harness,
  author       = {Gao, Leo and Tow, Jonathan and Abbasi, Baber and Biderman, Stella and Black, Sid and DiPofi, Anthony and Foster, Charles and Golding, Laurence and Hsu, Jeffrey and Le Noac'h, Alain and Li, Haonan and McDonell, Kyle and Muennighoff, Niklas and Ociepa, Chris and Phang, Jason and Reynolds, Laria and Schoelkopf, Hailey and Skowron, Aviya and Sutawika, Lintang and Tang, Eric and Thite, Anish and Wang, Ben and Wang, Kevin and Zou, Andy},
  title        = {A framework for few-shot language model evaluation},
  month        = sep,
  year         = 2021,
  publisher    = {Zenodo},
  version      = {v0.0.1},
  doi          = {10.5281/zenodo.5371628},
  url          = {https://doi.org/10.5281/zenodo.5371628}
}

@article{zhang2023h2o,
  title={{H2O}: Heavy-hitter oracle for efficient generative inference of large language models},
  author={Zhang, Zhenyu and Sheng, Ying and Zhou, Tianyi and Chen, Tianlong and Zheng, Lianmin and Cai, Ruisi and Song, Zhao and Tian, Yuandong and R{\'e}, Christopher and Barrett, Clark and others},
  journal={Advances in Neural Information Processing Systems},
  volume={36},
  pages={34661--34710},
  year={2023}
}

@article{xiao2023efficient,
  title={Efficient streaming language models with attention sinks},
  author={Xiao, Guangxuan and Tian, Yuandong and Chen, Beidi and Han, Song and Lewis, Mike},
  journal={arXiv preprint arXiv:2309.17453},
  year={2023}
}

@inproceedings{ainslie2023gqa,
  title={{GQA}: Training generalized multi-query transformer models from multi-head checkpoints},
  author={Ainslie, Joshua and Lee-Thorp, James and De Jong, Michiel and Zemlyanskiy, Yury and Lebr{\'o}n, Federico and Sanghai, Sumit},
  booktitle={Proceedings of the 2023 Conference on Empirical Methods in Natural Language Processing},
  pages={4895--4901},
  year={2023}
}

@article{liu2024deepseek,
  title={{Deepseek-V2}: A strong, economical, and efficient mixture-of-experts language model},
  author={Liu, Aixin and Feng, Bei and Wang, Bin and Wang, Bingxuan and Liu, Bo and Zhao, Chenggang and Dengr, Chengqi and Ruan, Chong and Dai, Damai and Guo, Daya and others},
  journal={arXiv preprint arXiv:2405.04434},
  year={2024}
}

@article{vaswani2017attention,
  title={Attention is all you need},
  author={Vaswani, Ashish and Shazeer, Noam and Parmar, Niki and Uszkoreit, Jakob and Jones, Llion and Gomez, Aidan N and Kaiser, {\L}ukasz and Polosukhin, Illia},
  journal={Advances in neural information processing systems},
  volume={30},
  year={2017}
}

@article{shazeer2019fast,
  title={Fast transformer decoding: One write-head is all you need},
  author={Shazeer, Noam},
  journal={arXiv preprint arXiv:1911.02150},
  year={2019}
}

@article{duanmu2024skvq,
  title={{SKVQ}: Sliding-window key and value cache quantization for large language models},
  author={Duanmu, Haojie and Yuan, Zhihang and Li, Xiuhong and Duan, Jiangfei and Zhang, Xingcheng and Lin, Dahua},
  journal={arXiv preprint arXiv:2405.06219},
  year={2024}
}

@article{beltagy2020longformer,
  title={Longformer: The long-document transformer},
  author={Beltagy, Iz and Peters, Matthew E and Cohan, Arman},
  journal={arXiv preprint arXiv:2004.05150},
  year={2020}
}

@article{olmo20242olmo2furious,
      title={{2 OLMo 2 Furious}}, 
      author={Team OLMo and Pete Walsh and Luca Soldaini and Dirk Groeneveld and Kyle Lo and Shane Arora and Akshita Bhagia and Yuling Gu and Shengyi Huang and Matt Jordan and Nathan Lambert and Dustin Schwenk and Oyvind Tafjord and Taira Anderson and David Atkinson and Faeze Brahman and Christopher Clark and Pradeep Dasigi and Nouha Dziri and Michal Guerquin and Hamish Ivison and Pang Wei Koh and Jiacheng Liu and Saumya Malik and William Merrill and Lester James V. Miranda and Jacob Morrison and Tyler Murray and Crystal Nam and Valentina Pyatkin and Aman Rangapur and Michael Schmitz and Sam Skjonsberg and David Wadden and Christopher Wilhelm and Michael Wilson and Luke Zettlemoyer and Ali Farhadi and Noah A. Smith and Hannaneh Hajishirzi},
      year={2024},
      eprint={2501.00656},
      archivePrefix={arXiv},
      primaryClass={cs.CL},
      url={https://arxiv.org/abs/2501.00656}, 
}

@misc{lozhkov2024fineweb-edu,
    author       = { Lozhkov, Anton and Ben Allal, Loubna and von Werra, Leandro and Wolf, Thomas },  
    title        = {{FineWeb-Edu}: the Finest Collection of Educational Content }, 
    year         = 2024,  
    url          = { https://huggingface.co/datasets/HuggingFaceFW/fineweb-edu },  
    doi          = { 10.57967/hf/2497 },
    publisher    = { Hugging Face }
}

@misc{merity2016pointer,
      title={Pointer Sentinel Mixture Models},
      author={Stephen Merity and Caiming Xiong and James Bradbury and Richard Socher},
      year={2016},
      eprint={1609.07843},
      archivePrefix={arXiv},
      primaryClass={cs.CL}
}

@article{raffel2020exploring,
  title={Exploring the limits of transfer learning with a unified text-to-text transformer},
  author={Raffel, Colin and Shazeer, Noam and Roberts, Adam and Lee, Katherine and Narang, Sharan and Matena, Michael and Zhou, Yanqi and Li, Wei and Liu, Peter J},
  journal={Journal of machine learning research},
  volume={21},
  number={140},
  pages={1--67},
  year={2020}
}

@inproceedings{bai2024longbench,
    title = "{L}ong{B}ench: A Bilingual, Multitask Benchmark for Long Context Understanding",
    author = "Bai, Yushi and Lv, Xin  and Zhang, Jiajie  and Lyu, Hongchang  and
      Tang, Jiankai  and Huang, Zhidian  and Du, Zhengxiao  and Liu, Xiao  and Zeng, Aohan  and Hou, Lei  and Dong, Yuxiao  and Tang, Jie  and Li, Juanzi",
    booktitle = "Proceedings of the 62nd Annual Meeting of the Association for Computational Linguistics (Volume 1: Long Papers)",
    month = aug,
    year = "2024",
    address = "Bangkok, Thailand",
    publisher = "Association for Computational Linguistics",
    url = "https://aclanthology.org/2024.acl-long.172",
    doi = "10.18653/v1/2024.acl-long.172",
    pages = "3119--3137",
}

@article{hendryckstest2021,
  title={Measuring Massive Multitask Language Understanding},
  author={Dan Hendrycks and Collin Burns and Steven Basart and Andy Zou and Mantas Mazeika and Dawn Song and Jacob Steinhardt},
  journal={Proceedings of the International Conference on Learning Representations (ICLR)},
  year={2021}
}

@inproceedings{zellers2019hellaswag,
    title={{HellaSwag}: Can a Machine Really Finish Your Sentence?},
    author={Zellers, Rowan and Holtzman, Ari and Bisk, Yonatan and Farhadi, Ali and Choi, Yejin},
    booktitle ={Proceedings of the 57th Annual Meeting of the Association for Computational Linguistics},
    year={2019}
}

@inproceedings{talmor-etal-2019-commonsenseqa,
    title = "{C}ommonsense{QA}: A Question Answering Challenge Targeting Commonsense Knowledge",
    author = "Talmor, Alon  and
      Herzig, Jonathan  and
      Lourie, Nicholas  and
      Berant, Jonathan",
    booktitle = "Proceedings of the 2019 Conference of the North {A}merican Chapter of the Association for Computational Linguistics: Human Language Technologies, Volume 1 (Long and Short Papers)",
    month = jun,
    year = "2019",
    address = "Minneapolis, Minnesota",
    publisher = "Association for Computational Linguistics",
    url = "https://aclanthology.org/N19-1421",
    doi = "10.18653/v1/N19-1421",
    pages = "4149--4158",
    archivePrefix = "arXiv",
    eprint        = "1811.00937",
    primaryClass  = "cs",
}

@misc{cobbe2021training,
      title={Training Verifiers to Solve Math Word Problems},
      author={Karl Cobbe and Vineet Kosaraju and Mohammad Bavarian and Jacob Hilton and Reiichiro Nakano and Christopher Hesse and John Schulman},
      year={2021},
      eprint={2110.14168},
      archivePrefix={arXiv},
      primaryClass={cs.LG}
}

@article{austin2021program,
  title={Program synthesis with large language models},
  author={Austin, Jacob and Odena, Augustus and Nye, Maxwell and Bosma, Maarten and Michalewski, Henryk and Dohan, David and Jiang, Ellen and Cai, Carrie and Terry, Michael and Le, Quoc and others},
  journal={arXiv preprint arXiv:2108.07732},
  year={2021}
}

@article{kusupati2022matryoshka,
  title={Matryoshka representation learning},
  author={Kusupati, Aditya and Bhatt, Gantavya and Rege, Aniket and Wallingford, Matthew and Sinha, Aditya and Ramanujan, Vivek and Howard-Snyder, William and Chen, Kaifeng and Kakade, Sham and Jain, Prateek and others},
  journal={Advances in Neural Information Processing Systems},
  volume={35},
  pages={30233--30249},
  year={2022}
}

@article{lin2024matryoshkakv,
  title={{MatryoshkaKV}: Adaptive kv compression via trainable orthogonal projection},
  author={Lin, Bokai and Zeng, Zihao and Xiao, Zipeng and Kou, Siqi and Hou, Tianqi and Gao, Xiaofeng and Zhang, Hao and Deng, Zhijie},
  journal={arXiv preprint arXiv:2410.14731},
  year={2024}
}

@article{lambert2024tulu,
  title={Tulu 3: Pushing frontiers in open language model post-training},
  author={Lambert, Nathan and Morrison, Jacob and Pyatkin, Valentina and Huang, Shengyi and Ivison, Hamish and Brahman, Faeze and Miranda, Lester James V and Liu, Alisa and Dziri, Nouha and Lyu, Shane and others},
  journal={arXiv preprint arXiv:2411.15124},
  year={2024}
}

@article{meng2025transmla,
  title={Transmla: Multi-head latent attention is all you need},
  author={Meng, Fanxu and Tang, Pingzhi and Tang, Xiaojuan and Yao, Zengwei and Sun, Xing and Zhang, Muhan},
  journal={arXiv preprint arXiv:2502.07864},
  year={2025}
}

@article{xu2024perfect,
  title={The perfect blend: Redefining rlhf with mixture of judges},
  author={Xu, Tengyu and Helenowski, Eryk and Sankararaman, Karthik Abinav and Jin, Di and Peng, Kaiyan and Han, Eric and Nie, Shaoliang and Zhu, Chen and Zhang, Hejia and Zhou, Wenxuan and others},
  journal={arXiv preprint arXiv:2409.20370},
  year={2024}
}

@inproceedings{yuan2025native,
  title={Native sparse attention: Hardware-aligned and natively trainable sparse attention},
  author={Yuan, Jingyang and Gao, Huazuo and Dai, Damai and Luo, Junyu and Zhao, Liang and Zhang, Zhengyan and Xie, Zhenda and Wei, Yuxing and Wang, Lean and Xiao, Zhiping and others},
  booktitle={Proceedings of the 63rd Annual Meeting of the Association for Computational Linguistics (Volume 1: Long Papers)},
  pages={23078--23097},
  year={2025}
}

@misc{openai2025gptoss,
  author       = {{OpenAI}},
  title        = {Introducing gpt-oss},
  year         = {2025},
  month        = aug,
  howpublished = {\url{https://openai.com/index/introducing-gpt-oss/}},
  note         = {Accessed: 2026-10-01}
}

@inproceedings{kai2026freqkv,
  title={FreqKV: key-value compression in frequency domain for context window extension},
  author={Kai, Jushi and Wang, Yixuan and Zeng, Boyi and Bai, Haoli and Jiang, Bo and He, Ziwei and Lin, Zhouhan},
  booktitle={International Conference on Learning Representations},
  volume={2026},
  pages={105568--105587},
  year={2026}
}

\appendix

\section{Details of DAR-opt}
\label{app:dar-opt}

This section provides the detailed formulation of DAR-opt, including the decoupled content and rotary components and weight absorption. \nop{For clarity, we present the content-side derivation for a single attention head; the multi-head implementation follows the same formulation independently for each head.}

\subsection{Content and Rotary Representations}

Given a hidden state $\mathbf{h}_j\in\mathbb{R}^d$, DAR-opt constructs two content latents:
\begin{equation}
\mathbf{c}^L_j = \mathbf{h}_j \mathbf{W}^{L}_{\mathrm{down}}, \qquad
\mathbf{c}^D_j = \mathbf{h}_j \mathbf{W}^{D}_{\mathrm{down}},
\end{equation}
where
\[
\mathbf{W}_{\mathrm{down}}^L\in\mathbb{R}^{d\times d_c^L},
\qquad
\mathbf{W}_{\mathrm{down}}^D\in\mathbb{R}^{d\times d_c^D},
\]
and $d_c^L>d_c^D$. The local path uses $\mathbf{c}_j^L$, while the distant path uses the lower-dimensional $\mathbf{c}_j^D$.

Following decoupled RoPE, positional information is represented separately. The rotary key is
\begin{equation}
    \mathbf{k}_j^R
    =
    \operatorname{RoPE}\!\left(\mathbf{h}_j\mathbf{W}_{K}^{R}\right)
    \in\mathbb{R}^{d_r},
\end{equation}
where $d_r$ denotes the rotary dimension. The rotary key is shared across attention heads as in MLA~\cite{liu2024deepseek}. For attention head $a$, the content and rotary queries are computed directly from the hidden state:
\begin{equation}
\label{eq:q_c_and_q_r}
\mathbf{q}^C_{i,a} = \mathbf{h}_i \mathbf{W}^{C}_{Q,a}, \qquad \mathbf{q}^R_{i,a} = \mathrm{RoPE}\!\left(\mathbf{h}_i \mathbf{W}^{R}_{Q,a}\right),
\end{equation}
where $\mathbf{W}^{C}_{Q,a} \in \mathbb{R}^{d \times d_h}$,
$\mathbf{W}^{R}_{K}, \mathbf{W}^{R}_{Q,a} \in \mathbb{R}^{d \times d_r}$,
and $d_h$ denotes the per-head content dimension.
Thus, DAR-opt does not introduce a compressed query latent; distance-adaptive compression is applied only to the cached KV content representations.

\subsection{Content Attention and Weight Absorption}

We first describe the local path for attention head $a$. The content key reconstructed from the local latent is
\begin{equation}
    \mathbf{k}_{j,a}^{C,L}
    =
    \mathbf{c}_j^L \mathbf{W}_{K,a}^L,
\end{equation}
where $\mathbf{W}_{K,a}^L\in\mathbb{R}^{d_c^L\times d_h}$, and the content query $\mathbf{q}^C_{i,a}$ is given by Eq.~\eqref{eq:q_c_and_q_r}.
Here $\mathbf{W}^{L}_{K,a}$ (and $\mathbf{W}^{L}_{V,a}$ below) plays the role of the up-projection: unlike the reference implementation in Section~\ref{sec:hybird_attn}, which first maps the bottlenecked representation back to dimension $d$ with $\mathbf{W}_{\mathrm{up}}$ and then applies $\mathbf{W}_K$ and $\mathbf{W}_V$, DAR-opt merges the two into a single projection per path, corresponding to $\mathbf{W}^{UK}$ and $\mathbf{W}^{UV}$ in MLA.
The content contribution to the attention score can be written as
\begin{align}
    \mathbf{q}_{i,a}^{C}
    \left(\mathbf{k}_{j,a}^{C,L}\right)^\top
    &=
    \mathbf{h}_i\mathbf{W}_{Q,a}^{C}
    \left(\mathbf{c}_j^L\mathbf{W}_{K,a}^{L}\right)^\top \\
    &=
    \mathbf{h}_i
    \mathbf{W}_{Q,a}^{C}
    \left(\mathbf{W}_{K,a}^{L}\right)^\top
    \left(\mathbf{c}_j^L\right)^\top .
\end{align}
We therefore define the absorbed query projection
\begin{equation}
    \widetilde{\mathbf{W}}_{Q,a}^{C,L}
    =
    \mathbf{W}_{Q,a}^{C}
    \left(\mathbf{W}_{K,a}^{L}\right)^\top,
\end{equation}
so that
\begin{equation}
    \mathbf{q}_{i,a}^{C}
    \left(\mathbf{k}_{j,a}^{C,L}\right)^\top
    =
    \mathbf{h}_i
    \widetilde{\mathbf{W}}_{Q,a}^{C,L}
    \left(\mathbf{c}_j^L\right)^\top.
\end{equation}
Thus, the full content key does not need to be materialized during decoding.

The distant path reconstructs its content key from the distant latent,
$\mathbf{k}^{C,D}_{j,a} = \mathbf{c}^D_j \mathbf{W}^{D}_{K,a}$ with $\mathbf{W}^{D}_{K,a} \in \mathbb{R}^{d^D_c \times d_h}$, and is absorbed analogously using
\begin{equation}
    \widetilde{\mathbf{W}}_{Q,a}^{C,D}
    =
    \mathbf{W}_{Q,a}^{C}
    \left(\mathbf{W}_{K,a}^{D}\right)^\top,
\end{equation}
which gives
\begin{equation}
    \mathbf{q}_{i,a}^{C}
    \left(\mathbf{k}_{j,a}^{C,D}\right)^\top
    =
    \mathbf{h}_i
    \widetilde{\mathbf{W}}_{Q,a}^{C,D}
    \left(\mathbf{c}_j^D\right)^\top.
\end{equation}

\subsection{Attention Score}

For attention head $a$, the attention score combines the content and rotary terms:
\begin{equation}
s_{ij,a}
=
\alpha
\begin{cases}
\mathbf{h}_i
\widetilde{\mathbf{W}}_{Q,a}^{C,L}
(\mathbf{c}_j^L)^\top
+
\mathbf{q}_{i,a}^R(\mathbf{k}_j^R)^\top,
& i-j<w, \\[4pt]
\mathbf{h}_i
\widetilde{\mathbf{W}}_{Q,a}^{C,D}
(\mathbf{c}_j^D)^\top
+
\mathbf{q}_{i,a}^R(\mathbf{k}_j^R)^\top,
& \text{otherwise}.
\end{cases}
\end{equation}
Following MLA~\cite{liu2024deepseek}, the scores are scaled by $\alpha = 1/\sqrt{d_h + d_r}$. Local and distant scores are normalized together using a single Softmax over all preceding positions, giving attention weights $p_{ij,a}$. The rotary component is shared across the two paths; DAR-opt changes only the dimensionality of the cached content latent according to token distance.

\subsection{Value-Side Weight Absorption}

The values are reconstructed from the latents in the same way,
\begin{equation}
    \mathbf{v}_{j,a}^{L} = \mathbf{c}_j^L \mathbf{W}_{V,a}^{L}, \qquad
    \mathbf{v}_{j,a}^{D} = \mathbf{c}_j^D \mathbf{W}_{V,a}^{D},
\end{equation}
with $\mathbf{W}^{L}_{V,a} \in \mathbb{R}^{d^L_c \times d_h}$ and $\mathbf{W}^{D}_{V,a} \in \mathbb{R}^{d^D_c \times d_h}$. Let $\mathbf{W}_{O,a} \in \mathbb{R}^{d_h \times d}$ be the slice of the output projection for head $a$, and define
\begin{equation}
    \widetilde{\mathbf{W}}_{VO,a}^{L} = \mathbf{W}_{V,a}^{L}\mathbf{W}_{O,a}, \qquad
    \widetilde{\mathbf{W}}_{VO,a}^{D} = \mathbf{W}_{V,a}^{D}\mathbf{W}_{O,a}.
\end{equation}
The output of head $a$ after the output projection is then
\begin{equation}
    \mathbf{o}_{i,a}\mathbf{W}_{O,a}
    =
    \Big(\sum_{j:\, i-j<w} p_{ij,a}\, \mathbf{c}^L_j\Big) \widetilde{\mathbf{W}}^{L}_{VO,a}
    +
    \Big(\sum_{j:\, i-j\ge w} p_{ij,a}\, \mathbf{c}^D_j\Big) \widetilde{\mathbf{W}}^{D}_{VO,a}.
\end{equation}
The attention-weighted sums are computed directly over the cached latents, so neither path materializes full-dimensional value states during decoding.

During training, DAR-opt uses the unabsorbed form, reconstructing content keys and values from the latents; the absorbed matrices are computed once from the trained weights and used only for decoding, which leaves the model's outputs unchanged.
\section{Perplexity Evaluation Details}
\label{app:evaluation}

\begin{figure}[h]
  \centering
  \includegraphics[
    width=0.6\linewidth,
    trim={5 5 5 5},
    clip
  ]{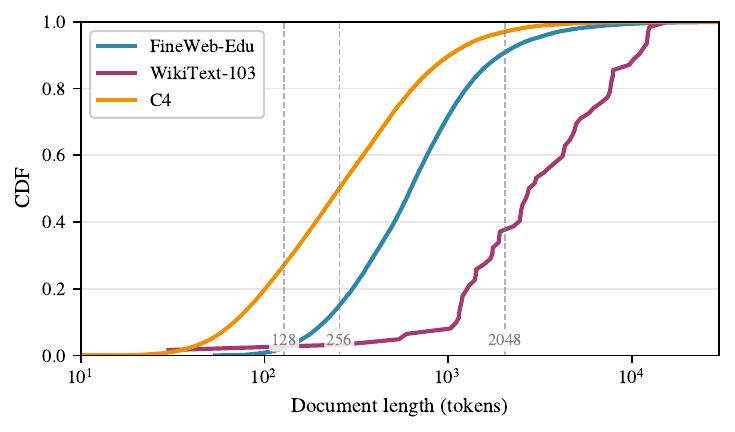}
  \caption{Document-length distribution (CDF) of three perplexity evaluation corpora, tokenized with the Pythia tokenizer. The vertical dashed lines mark the local window size ($w=128$) and the training sequence length ($2,048$). The distribution shows that the majority of evaluation tokens lie well beyond the $w=128$ window, showing that evaluation frequently exercises the distant context path.}
  \label{fig:seq_length_cdf}
\end{figure}

We evaluate pretraining models on subsets of FineWeb-Edu, C4, and WikiText-103 using the Pythia tokenizer. For documents longer than the maximum training sequence length of 2,048 tokens, we use a rolling evaluation strategy to cover the full document without discarding tokens. Relative perplexity is computed independently on each dataset with respect to the reference baseline specified for each table and then averaged across the three datasets. Figure~\ref{fig:seq_length_cdf} shows the document-length distributions of the three perplexity evaluation corpora. Most evaluation sequences extend beyond the local window size $w=128$, ensuring that evaluation substantially exercises the distant context path.

\section{Analysis across Token Distance}
\label{app:effective-dim}

We provide additional details for the query-conditioned analysis introduced in Section~\ref{sec:hypothesis_validation}.

A key or value state does not have a fixed token distance by itself, since its distance is defined relative to the query position. We therefore analyze the attention-weighted value information received by each query as a function of relative token distance.

For each query position $i$, we partition the preceding context into 29 consecutive distance bins of 64 tokens, covering relative distances 1--1,856. We stop at 1,856 rather than using the full 2,048-token context so that the farthest bin retains a sufficient number of samples. For each distance bin $b$, we compute
\begin{equation}
    \mathbf{m}_i^{(b)}
    =
    \sum_{j:\, i-j\in b} a_{ij}\mathbf{v}_j,
\end{equation}
where $a_{ij}$ is the attention weight from query position $i$ to token $j$. The resulting vector describes the attention-weighted value contribution from tokens within a particular distance range.

Each attention head is analyzed independently. For every distance bin, we collect 24,000 such vectors across queries. For each layer and head, we compute their covariance and apply PCA. We report $d_{90}$, the smallest number of principal components required to explain 90\% of the variance, normalized by the head dimension as $d_{90}/d_{\text{head}}$. A smaller value indicates that the variation among the attention-weighted value aggregates is concentrated in fewer directions. We average this value across all attention heads within each layer. We perform this analysis on the pretrained Pythia-2.8B-deduped checkpoint using FineWeb-Edu sequences.

This analysis characterizes query-conditioned, attention-weighted value aggregates and should not be interpreted as a direct measurement of the intrinsic rank of the raw key or value states.

\section{Training Trajectories}
\label{app:training-curves}

\begin{figure}[h]
    \centering
    \includegraphics[
        width=0.5\linewidth,
        trim={5 5 5 7},
        clip
    ]{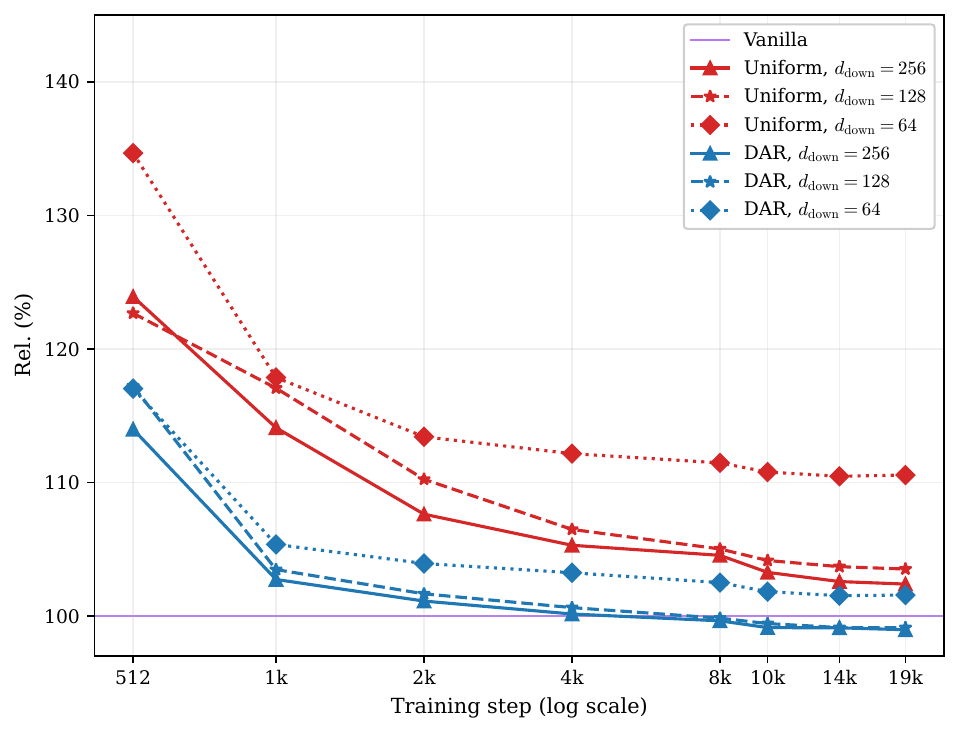}
    \caption{Average perplexity ratio relative to Vanilla (= 100\%, shown as horizontal line) across training steps at the Pythia-70M scale.}
    \label{fig:training_traj}
\end{figure}

Figure~\ref{fig:training_traj} shows the relative perplexity trajectory across pretraining at the Pythia-70M scale. DAR remains consistently closer to Vanilla than Uniform Reduction across bottleneck dimensions. By the end of training, DAR with $d_{\text{down}}\in\{128,256\}$ reaches performance comparable to Vanilla, while $d_{\text{down}}=64$ retains a modest gap. Uniform Reduction remains consistently worse across all three bottleneck dimensions.

\section{Window Size Ablation}
\label{app:window-ablation}

We study the effect of the local window size $w$ at the Pythia-160M scale with $d_{\text{down}} = 192$, while keeping all other training settings fixed. As shown in Table~\ref{tab:window}, the $w = 0$ setting, which corresponds to uniform dimensionality reduction, performs worst. Small windows ($w \in \{2, 4, 8\}$) perform similarly to one another, as do larger windows ($w \in \{32, 128, 512\}$), and the larger windows generally outperform the smaller ones.

\begin{table}[t]
    \centering
    \small
    \setlength{\tabcolsep}{6pt}
    \caption{Effect of the local window size $w$ at the Pythia-160M scale with $d_{\text{down}}=192$. The $w=0$ setting corresponds to uniform dimensionality reduction; $w=128$ is the default setting used in Table~\ref{tab:scaling_analysis}. Rel.(\%) is the average per-dataset perplexity ratio relative to Vanilla (lower is better).}
    \label{tab:window-ablation}
    \begin{tabular}{lc|ccc|c}
        \toprule
        \textbf{Model}
        & \textbf{$w$}
        & \textbf{FineWeb-Edu}
        & \textbf{C4}
        & \textbf{WikiText-103}
        & \textbf{Rel.(\%)} \\
        \midrule
        Vanilla
        & -- & 79.46 & 99.51 & 35.96 & 100.00 \\
        \midrule
        \multirow{7}{*}{DAR}
        & 0   & 81.81 & 100.66 & 36.92 & 102.26 \\
        & 2   & 80.46 & 98.47  & 36.51 & 100.58 \\
        & 4   & 80.28 & 98.59  & 36.36 & 100.40 \\
        & 8   & 80.26 & 98.59  & 36.61 & 100.62 \\
        & 32  & 78.35 & 96.70  & 35.99 & 98.61 \\
        & 128 & 78.93 & 97.84  & 36.05 & 99.30 \\
        & 512 & 79.07 & 97.33  & 36.43 & 99.53 \\
        \bottomrule
    \end{tabular}
    \label{tab:window}
\end{table}

\section{Extension to MLA-based LLaMA-3-8B}
\label{app:dar-mla}

We further evaluate DAR-opt at the 8B scale using a distillation setup. Following TransMLA~\cite{meng2025transmla}, we convert LLaMA-3-8B to an MLA-based architecture with content latent dimension $d_c=512$ and rotary dimension $d_r=64$. The converted MLA-512 model is kept fixed and used as both the teacher and reference model.

Rather than retraining the full 8B model, we use distillation to keep the computational cost manageable while testing whether DAR remains effective at a larger scale. Since MLA-512 serves as the teacher, the purpose of this experiment is not to outperform it, but to compare how well different compression schemes preserve its downstream performance, including long-context performance.

All compressed variants are trained by distilling the teacher's final hidden states with an L2 loss on Open-PerfectBlend~\cite{xu2024perfect}. Only the content components of the KV latent projections are trained; the rotary key and all other parameters are frozen and shared with the teacher. DAR-opt uses $d_c^L=512$ for the most recent $w=256$ tokens and $d_c^D=256$ for distant tokens. Both projections are initialized from the teacher, with the distant projection truncated to the first $d_c^D$ dimensions. The uniform MLA-272 and MLA-256 baselines are initialized using the same truncation and trained with the same objective, data, and hyperparameters. At an 8K context length, DAR-opt has an average cache size of $d_c^D+d_r+\frac{w d_c^L}{8192}=336$ dimensions per token.

We evaluate MMLU and GSM8K in the 5-shot setting and RULER at a context length of 8,192. As shown in Table~\ref{tab:dar-mla}, DAR-opt retains 97.29\% of the MLA-512 performance on average across the three benchmarks, including 97.78\% on RULER. At the same average cache size, MLA-272 retains 85.05\%, showing the advantage of distance-adaptive allocation over uniform compression. MLA-256 uses the same $d_c=256$ as the distant latent of DAR-opt but retains 84.01\%, further showing the benefit of preserving the high-dimensional content representation for local tokens.

\begin{table}[t]
    \centering
    \small
    \setlength{\tabcolsep}{4pt}
    \caption{Distillation-based evaluation on an MLA-converted LLaMA-3-8B model. MLA-512 serves as the reference and teacher model, and all compressed variants are trained under the same distillation setting. DAR-opt uses $w=256$, while MLA-272 provides a matched-cache baseline with the same average cache size. MLA-256 uses the same content dimension as the distant latent of DAR-opt. Avg.\ cache size is measured per token at 8K context. Rel.(\%) is the average of task-specific scores normalized by the corresponding MLA-512 scores.}
    \label{tab:dar-mla}
    \begin{tabular}{lccc|c|cccc}
        \toprule
        \textbf{Method}
        & \multicolumn{3}{c|}{\textbf{Latent configuration}}
        & \textbf{Avg.\ cache}
        & \multicolumn{4}{c}{\textbf{Downstream performance}} \\
        \cmidrule(lr){2-4}
        \cmidrule(lr){6-9}
        &
        \textbf{$d_c^L$}
        & \textbf{$d_c^D$}
        & \textbf{$d_r$}
        & \textbf{size}
        & \textbf{MMLU}
        & \textbf{GSM8K}
        & \textbf{RULER@8K}
        & \textbf{Rel.(\%)} \\
        \midrule
        MLA-512
        & 512 & 512 & 64
        & 576
        & 0.6221 & 0.7710 & 0.9495 & 100.00 \\
        MLA-272
        & 272 & 272 & 64
        & 336
        & 0.4826 & 0.6869 & 0.8402 & 85.05 \\
        MLA-256
        & 256 & 256 & 64
        & 320
        & 0.4673 & 0.6816 & 0.8405 & 84.01 \\
        \midrule
        \textbf{DAR-opt}
        & \textbf{512} & \textbf{256} & \textbf{64}
        & \textbf{336}
        & \textbf{0.5993} & \textbf{0.7536} & \textbf{0.9284} & \textbf{97.29} \\
        \bottomrule
    \end{tabular}
\end{table}


\end{document}